\documentclass[11pt,a4paper]{article}
\usepackage[hyperref]{emnlp-ijcnlp-2019}
\usepackage{times}
\usepackage{latexsym}
\usepackage{amssymb}
\usepackage{url}
\usepackage{examples}
\usepackage{multirow}
\usepackage{color}
\usepackage{xcolor,colortbl}
\usepackage{graphicx}
\usepackage{amsmath}
\definecolor{lightgrey}{RGB}{227,226,219}

\aclfinalcopy

\title{
Fine-grained Information Status Classification Using Discourse Context-Aware Self-Attention 
}

 \author{
Yufang Hou \\
 IBM Research Ireland, Dublin, Ireland\\ {\tt yhou@ie.ibm.com}\\
    }

\date{}

\begin{document}
\maketitle

\begin{abstract}
Previous work on bridging anaphora recognition \cite{houyufang13b} 
casts the problem as a subtask of learning fine-grained information status (IS). However, 
these systems heavily depend on many hand-crafted linguistic features. In this paper,
we propose a discourse context-aware self-attention neural network model for fine-grained IS classification. On the ISNotes corpus \cite{markert12}, our model with the contextually-encoded word
representations (BERT) \cite{devlin2018bert} achieves new state-of-the-art performances on fine-grained IS classification, obtaining a 4.1\% absolute overall accuracy improvement compared to \newcite{houyufang13b}. More importantly, we also show an improvement 
of 3.9\% F1 for 
bridging anaphora recognition without using any complex hand-crafted semantic features designed for 
capturing the bridging phenomenon.
\end{abstract}

 \section{Introduction}
\label{sec:intro}
\emph{Information Structure} \cite{halliday67,prince81a,prince92,gundel93,lambrecht94,birner98,kruijff-korbayova03}
studies structural and semantic properties of a sentence according to its relation to the 
discourse context. Information structure affects how discourse entities are referred to in a
 text, which
is known as \emph{Information Status} \cite{halliday67,prince81a,nissim04}.
Specifically, information status (IS henceforth) reflects the accessibility of a discourse entity
based on the evolving discourse context and the speaker's assumption about the hearer's knowledge and beliefs.
For instance, according to \newcite{markert12}, \emph{old} mentions\footnote{A mention is a noun phrase which refers to a discourse entity and carries information status.} refer to entities that %are known to the hearer and
have been referred to previously; \emph{mediated} mentions have not
been mentioned before but are accessible to the hearer by reference to another \emph{old} mention or to prior world knowledge; and \emph{new} mentions refer to entities that are introduced to the discourse for the first time and are not known to the hearer before.

In this paper, we follow the IS scheme proposed by \newcite{markert12} and focus on learning fine-grained IS on written texts.  %When assigning IS for a mention, apart from the semantic/structure properties of the mention and its relation to the discourse context both play a role. 
A mention's semantic and syntactic properties can signal its information status. For instance, indefinite NPs tend to be \emph{new} and pronouns are likely to be \emph{old}. 
%Moreover, discourse context also plays a role when assigning IS for a mention. 
 Moreover, referential patterns of how a mention is referred to in a sentence also affect this mention's IS. %information status.
In Example \ref{exa:exa11},  
``Friends'' is a bridging anaphor even if we do not know the antecedent (i.e., \emph{she}); while the information status for ``Friends'' in Example \ref{exa:exa12} is \emph{mediated/worldKnowledge}. Section \ref{subsec:motivation} analyzes the characteristics of each IS category and %how they relate to the discourse context. 
the relations between IS and discourse context.

\begin{examples}
\item \label{exa:exa11} \emph{She} made money, but spent more. {\bfseries Friends} pitched in.
\item \label{exa:exa12} \underline{Friends} are part of the glue that holds life and faith together.
%\item \label{exa:exa1} \textbf{Repairs} are expected to cost about \$240 million.
%\item \label{exa:exa2} Also, most of the ramps have been closed for \textbf{repairs}.
\end{examples}

\begin{table*}[ht]
\begin{footnotesize}
\begin{tabular} {|l|l|l|c|c|c|}
\hline
&&& \multicolumn{3}{c|}{Factors affecting IS} \\
 &Description  & Example & Mention  & Local  & Previous \\
 &  &  & Properties &Context & Context\\
  \hline
 {old} &coreferent with an already introduced entity&\emph{he}, \emph{the president}& \checkmark &\checkmark &\checkmark \\ \hline
 {m/worldKnow.} &generally known to the hearer&\emph{Francis}, \emph{the pope}&\checkmark &\checkmark &\\ \hline
 {m/syntactic} &syntactically linked to other  \emph{old} or \emph{mediated} &\emph{their father}&\checkmark &&\\ 
 &mentions&\emph{a war in Africa}& &&\\ \hline
 {m/aggregate} &coordinated NPs where at least one element 
  &\emph{U.S. and Canada}&\checkmark & &\\ 
 &is \emph{old} or \emph{mediated} &\emph{he and his son}& & &\\ \hline
 {m/function} &refer to a value of a previously explicitly  &(the price went&\checkmark &\checkmark &\\ 
 &mentioned rise/fall function&down) \emph{6 cents}& & &\\ \hline
 {m/comparative} &usually contain a premodifier to indicate that &\emph{another law}&\checkmark &\checkmark & \checkmark \\ 
  & this entity is compared to another entity &\emph{further attacks}& & &  \\ \hline
 {m/bridging} & associative anaphors which link to previously &\emph{the price}&\checkmark &\checkmark & \checkmark \\ 
 & introduced related entities/events&\emph{the reason}& & & \\ \hline
 {new} &introduced into the discourse for the first time &\emph{a reader}&\checkmark &\checkmark & \\  
&and not known to the hearer before&\emph{politics}&&& \\  \hline
\end{tabular}
\end{footnotesize}
\caption{Information status categories and their main affecting factors. ``Local context'' means 
the sentence $s$ which contains the target mention, ``Previous context'' indicates all sentences from the discourse which occur before $s$.}
\label{tab:motivation}
\end{table*}

In this work, we propose a discourse context-aware self-attention neural network model for fine-grained IS classification. We find that the sentence containing the target mention as well as the lexical overlap information between the target mention and the preceding mentions are the most important discourse context when assigning IS for a mention. With self-attention, our model can capture important signals within a mention and the interactions between the mention and its context. On the ISNotes corpus \cite{markert12}, our model with the contextually-encoded word
representations (BERT) \cite{devlin2018bert} achieves new state-of-the-art performances on fine-grained IS classification, obtaining a 4.1\% absolute overall accuracy improvement compared to \newcite{houyufang13b}. More importantly, we also show an improvement 
of 3.9\% F1 for 
bridging anaphora recognition without using any sophisticated hand-crafted semantic features.
%\footnote{The experimental data and code are publicly available at: \url{http://anonymized-for-blind-review}} 

\section{Related Work}
\label{sec:relatedwork}
\paragraph{IS classification and bridging anaphora recognition.}
Bridging resolution \cite{houyufang14,houyufang18c} contains two sub tasks: identifying bridging 
anaphors \cite{markert12,houyufang13b,houyufang16} and finding the correct antecedent among candidates \cite{houyufang13a,houyufang18b,houyufang18}. Previous work handle bridging anaphora recognition as part of IS classification problem.
\newcite{markert12} applied joint inference for IS classification on the ISNotes corpus but reported very low results on bridging recognition.
Built on this work, \newcite{houyufang13b} designed many linguistic features to capture bridging and integrated them into a cascading collective classification algorithm. Differently, \newcite{houyufang16} used an attention-based LSTM model based on GloVe vectors and a small set of features for IS classification. The author reported similar results as \newcite{houyufang13b} regarding the overall IS classificiation accuracy but the result on bridging anaphora recognition is much worse than \newcite{houyufang13b}.  
%combines GloVe word embeddings with a small set of features  achieved  competitive  results  on  ISNotes  compared  to  our
%collective classification approach (Hou 2016)

\newcite{rahman12} incorporated carefully designed rules into an SVM$^{multiclass}$ algorithm for IS classification on the Switchboard dialogue corpus \cite{nissim04}. \newcite{cahill12} trained a CRF model with syntactic and surface features for fine-grained IS classification on the German DIRNDL radio news corpus \cite{riester10}\footnote{Bridging antecedent information was not annotated in Switchboard and DIRNDL. Also IS annotation in Switchboard includes non-anaphoric cases.}.

Different from the above mentioned work, we do not use any complicated hand-crafted features and our model improves the previous state-of-the-art results 
on both overall IS classification accuracy and bridging recognition 
by a large margin on the ISNotes corpus.

\paragraph{Self-attention.} 
Recently, multi-head self-attention encoder \cite{ashish17} has been shown to perform well in various NLP tasks, including semantic role labelling \cite{strubell18}, question answering and natural language inference \cite{devlin2018bert}. In our model, we create
a ``pseudo sentence'' for each mention and apply the transformer encoder for our task.
%by concatenating the mention with its discourse context using a special delimiter. Each pseudo sentence starts with an unique IS prediction token ``[IS]''. 
The self-attention mechanism allows our model to attend to both the context and the mention itself for clues which are helpful for predicting the mention's IS.
%When the model processes each word in the sentence, self-attention allows it to attend to other positions of the sequence for clues which can lead to a better encoding for this word. 

\paragraph{Fine-tuning with contextual word embeddings.}
Recent work \cite{peter18,howard18,devlin2018bert} have shown that a range of downstream NLP tasks benefit from fine-tuning task-specific parameters with pre-trained contextual word representations. Our work belongs to this category and we fine-tune our model based on BERT$_{BASE}$ representations \cite{devlin2018bert}.

\section{Approach}
\label{sec:method}

\subsection{Information Status and Discourse Context}
\label{subsec:motivation}
The IS scheme proposed by \newcite{markert12} adopts  three major IS categories (\emph{old}, \emph{new}
and \emph{mediated}) from \newcite{nissim04} and distinguishes six subcategories for \emph{mediated}. Table \ref{tab:motivation} lists the definitions for these IS categories
and summarizes the main affecting factors for each IS class.
%poias well as the relations between each IS category and its affecting factors.  

\begin{figure*}[t]
\begin{center}
\includegraphics[width=0.79\textwidth]{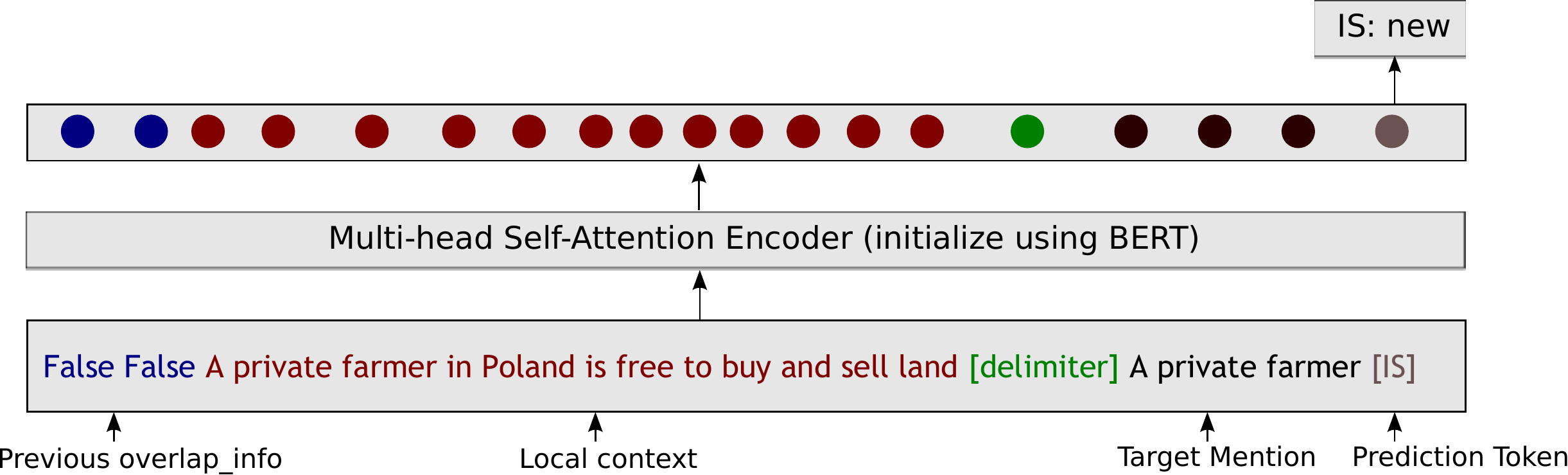}
\end{center}
\caption{Fine-grained IS classification with discourse context-aware self-attention.}
\label{fig:model}
\end{figure*}
%\vspace{-0.5cm}

%There are three factors which are related 
%There are three factors which influence the IS of a mention. 
As described in Section \ref{sec:intro}, a mention's internal syntactic and semantic properties can signal its IS. For instance, a mention containing
a possessive pronoun modifier is likely to be \emph{mediated/syntactic} (e.g., \emph{their father}); and a \emph{mediated/comparative} mention often contains a premodifier indicating that this entity is compared to another preceding entity (e.g., \emph{further attacks}).

In addition, for some IS classes, the ``local context'' (the sentence $s$ which contains the target mention) and 
``previous context'' (sentences from the discourse which precede $s$) play an important role when assigning IS to a mention. Example \ref{exa:exa11} and Example \ref{exa:exa12} in Section \ref{sec:intro} demonstrate the role of the local context for IS. And sometimes we need to look at the previous context when deciding IS for a mention. In Example \ref{exa:exa4}, without looking at the previous context, we tend to think the IS for ``Poland'' in the second sentence is \emph{mediated/WorldKnowledge}. Here the correct IS for  ``Poland'' is \emph{old} because it is mentioned before in the previous context.

%In Example \ref{exa:exa3}, the local context indicates that ``6 cents'' is the value of the function ``went down'', therefore the IS for ``6 cents'' should be \emph{medicated/function}.

\begin{examples}
%\item \label{exa:exa3} In trading on the American Stock Exchange, Delmed went down \textbf{6 cents}. 
\item \label{exa:exa4} [Previous context:]  In \textbf{Poland}, only 4\% of all investment goes toward making things farmers want; in the West, it is closer to 20\%. \newline [Local context:]  A private farmer in \textbf{Poland} is free to buy and sell land.
\end{examples}

\subsection{IS Classification with Discourse Context-Aware Self-Attention}
\label{subsec:model}
To account for the different factors described in the previous section when predicting IS for a mention, we create a novel ``pseudo sentence'' for each mention and apply the  multi-head self-attention encoder \cite{ashish17} for this sentence. 
%This process casts the IS classification task as a sentence classification problem. 

Figure \ref{fig:model} depicts the high-level structure of our model. The pseudo sentence consists of five parts: previous overlap\_info, local context, the delimiter token ``[delimiter]'', the content of the target mention, and the IS prediction token ``[IS]''. The previous overlap\_info part contains two tokens, which indicate %the overlap information between the target mention and the mentions in the previous context (i.e., full string match and head match). And the local context is the sentence containing the target mention. 
whether the target mention has the same string/head with a mention from the preceding sentences. And the local context is the sentence containing the target mention. 

The final prediction is made based on the hidden state of the prediction token ``[IS]''. 
In principle, the structure of the pseudo sentence and the mechanism of multi-head self-attention help the model to learn the important cues from both the mention and its discourse context when predicting IS.

\subsection{Model Parameters}
\label{subsec:parameter}
Our context-aware self-attention model has 12 transformer blocks, 768 hidden units, and 12 self-attention heads. We first initialize our model using BERT$_{BASE}$, then fine-tune the model for 3 epochs with the learning rate of $5e-5$. During training and testing,  the max token size of the pseudo sentence is set as 128.

\section{Experiments}
\vspace{-5pt}
\subsection{Experimental Setup}
We perform experiments on the ISNotes corpus \cite{markert12}, which contains 10,980 mentions annotated for information status
in 50 news texts. 
%taken from the Wall Street Journal portion of the OntoNotes corpus \cite{ontonotes4.0data}. 
Table \ref{tab:is} shows the IS distribution in ISNotes.

\begin{table}[h]
\begin{center}
\begin{footnotesize}
\begin{tabular}{lrrr}
Mentions & &10,980& \\ \hline\hline
old & & 3237&29.5\% \\ \hline
mediated & & 3,708&33.8\%\\ \hline
& syntactic & 1,592&14.5\%\\
& world knowledge & 924&8.4\%\\
& bridging & 663&6.0\%\\
& comparative & 253&2.3\%\\ %how many are deictic?
& aggregate & 211&1.9\%\\
& func & 65&0.6\%\\
 \hline
new & & 4,035&36.7\%\\
\end{tabular}
\end{footnotesize}
\end{center}
\begin{footnotesize}
\caption{\label{tab:is}  IS distribution in ISNotes.}
\end{footnotesize}
\end{table}

\begin{table*}[t]
\begin{footnotesize}
\begin{tabular}{|p{1.8cm}|p{0.466cm}p{0.466cm}p{0.466cm}|p{0.466cm}p{0.466cm}p{0.466cm}|p{0.466cm}p{0.466cm}p{0.466cm}|p{0.466cm}p{0.466cm}p{0.466cm}|p{0.466cm}p{0.466cm}p{0.466cm}|}
\hline
 &  \multicolumn{3}{c|}{\emph{collective}} &\multicolumn{3}{c|}{\emph{cascade collective}} & \multicolumn{3}{c|}{\emph{self-attention }} & \multicolumn{3}{c|}{\emph{self-attention}}& \multicolumn{3}{c|}{\emph{self-attention}}\\ 
 &  \multicolumn{3}{c|}{Hou et al.(2013)} &\multicolumn{3}{c|}{Hou et al.(2013)} & \multicolumn{3}{c|}{\emph{wo context}} & \multicolumn{3}{c|}{\emph{with context I}}& \multicolumn{3}{c|}{\emph{with context II}}\\ 
 & R & P & F & R & P & F& R & P & F& R & P & F& R & P & F\\ \hline
\hline
 &&&&&&&&&&&&&&& \\
 {old} & 84.4&86.0&85.2&82.2&87.2&84.7&79.3&82.4&80.8&80.1&85.8&82.9&86.8&90.9&\textbf{88.8}\\
 {m/worldKnow.}&67.4&77.3&72.0&67.2&77.2&71.9&57.8&62.2&59.9&60.8&64.6&62.7&74.6&79.6&\textbf{77.0}\\
 {m/syntactic} & 82.2&81.9&82.0&81.6&82.5&82.0&83.7&81.7&82.7&83.2&82.6&82.9&83.6&83.0&\textbf{83.3}\\
 {m/aggregate}&64.5&79.5&71.2&63.5&77.9&70.0&77.7&75.2&76.5&76.3&79.7&78.0&76.8&80.6&\textbf{78.6} \\
 {m/function} & 67.7&72.1&\textbf{69.8}&67.7&72.1&\textbf{69.8}&42.2&56.3&48.2&54.7&61.4&57.9&50.0&74.4&59.8\\
 {m/comparative} & 81.8&82.1&82.0&86.6&78.2&82.2&89.3&88.6&89.0&90.1&86.4&88.2&91.7&89.2&\textbf{90.4}\\
 \rowcolor{lightgrey}
 {m/bridging}& 19.3&39.0&25.8&44.9&39.8&42.2&37.6&48.6&42.4&43.6&51.6&\textbf{47.3}&42.5&50.4&46.1\\
 {new} & 86.5&76.1&81.0&83.0&78.1&80.5&86.1&80.0&82.9&88.3&81.0&84.5&88.5&82.0&\textbf{85.1}\\
\hline
 acc &\multicolumn{3}{c|}{78.9} & \multicolumn{3}{c|}{78.6} & \multicolumn{3}{c|}{78.1} & \multicolumn{3}{c|}{79.8}&\multicolumn{3}{c|}{\textbf{83.0}}\\
\hline
\end{tabular}
\end{footnotesize}
\caption{Results of the discourse context-aware self-attention model compared to the baselines. Bolded scores indicate the best performance for each IS class. The improvements of \emph{self-attention with context I} and \emph{self-attention with context II} over the baselines are statistically significant at $p<0.01$ using randomization test.}
\label{tab:results}
\end{table*}

\vspace{-12pt}
Following \newcite{houyufang13b}, all experiments are performed via 10-fold cross-validation on
documents. We report overall accuracy as well as precision, recall and F-measure per IS class. In the following, we describe the baselines as well as our model with different settings.

\paragraph{\emph{collective} (baseline1).} \newcite{houyufang13b} applied collective classification to account for the linguistic relations among IS categories.
 They explored a wide range of features (34 in total), including a large number of lexico-semantic features (for recognizing bridging) as well as a couple of surface features and syntactic features.
 
\paragraph{\emph{cascaded collective} (baseline2).} This is the cascading minority preference system for bridging anaphora recognition from \newcite{houyufang13b}. 

\paragraph{\emph{self-attention wo context}.} We apply our model (see Section \ref{sec:method}) 
on the pseudo sentences containing only the target mentions and the prediction token ``[IS]''.

\paragraph{\emph{self-attention with context I}.} Based on \emph{self-attention wo context}, we add the local context in the pseudo sentences.

\paragraph{\emph{self-attention with context II}.} Based on \emph{self-attention with context I}, we add the previous overlap\_info part in the pseudo sentences.

\subsection{Results and Discussion}
Table \ref{tab:results} shows the results of our models compared to the baselines. 
Surprisingly, our model %with the pseudo sentences containing only the mentions
considering only the content of mentions 
 (\emph{self-attention wo context}) achieves competitive results as the baseline \emph{cascade collective} which explores many hand-crafted linguistic features. 
Also \emph{self-attention wo context} outperforms the two baselines on several IS categories (\emph{m/syntactic}, \emph{m/aggregate}, \emph{m/comparative}, \emph{m/bridging} and \emph{new}). In Section \ref{subsec:motivation}, we analyze that \emph{m/syntactic} and 
\emph{m/aggregate} are often signaled by mentions' internal syntactic structures, and that the semantics of certain premodifiers is a strong signal for \emph{m/comparative}. 
The improvements on these categories show that our model can capture the semantic/syntactic properties of a mention when predicting its IS.

The continuous improvements on \emph{self-attention with context I} and \emph{self-attention with context II} show the impact of the local context and the previous context on IS prediction, respectively. It seems that the local context has more impact on \emph{m/bridging} and \emph{new}, whereas the previous context has more impact on \emph{old} and \emph{m/worldKnowledge}.

For \emph{self-attention with context I}, we also tried to add the previous $k$ sentences  ($k=1$ and  $k=2$) into the current local context to see whether the broader local context can help us to capture bridging better. However, we found that the overall results on both settings are similar as the current one.

Overall, we achieve the new state-of-the-art results on bridging anaphora recognition with the local context model (\emph{self-attention with context I}). And our full model (\emph{self-attention with context II}) achieves an overall accuracy of 83\% on IS classification, obtaining a 4.1\% and 4.4\% absolute improvements over the two baselines (\emph{collective} and \emph{cascade collective}), respectively. Our full model beats the two strong baselines on most IS categories except \emph{m/function}. This is because
%\emph{m/function} is a very rare IS category with only 65 
 there are only 65 \emph{m/function} mentions in ISNotes. 
With such a small amount of training data, it is hard for our model to learn patterns for this category. 
 %Therefore it is hard for our model to learn patterns for this category with such a small amount of training instances.

\section{Conclusions}
We develop a discourse context-aware self-attention model for IS classification.
%We cast the IS classification problem as a sentence classification task by creating 
%a ``pseudo sentence'' for each mention. 
%We create a novel ``pseudo sentence'' for each mention 
 %by concatenating an unique IS prediction token ``[IS]'' with the mention and its discourse context. 
%Also a special delimiter is used to separate the mention with its context.
%The mention and the discourse context is separated using a special delimiter.
% Such design allows the model to capture both clues from the mention and its context %when predicting IS.
Our model does not contain any complex hand-crafted semantic features and achieves the new state-of-the-art results for IS classification and bridging anaphora recognition on ISNotes.

\bibliographystyle{acl_natbib}
\bibliography{../bib/lit/lit}

%%================================================================
\end{document}